\documentclass{article}

\usepackage{microtype}
\usepackage{graphicx}
\usepackage{subcaption}
\usepackage{tabularx}
\usepackage{array}
\usepackage{makecell}
\usepackage{stfloats}
\usepackage{booktabs} % for professional tables

\usepackage{hyperref}

\usepackage[accepted]{icml2026}

\usepackage{amsmath}
\usepackage{amssymb}
\usepackage{mathtools}
\usepackage{amsthm}

\usepackage[capitalize,noabbrev]{cleveref}

\theoremstyle{plain}

\theoremstyle{definition}

\theoremstyle{remark}

\usepackage[textsize=tiny]{todonotes}

\icmltitlerunning{A Precedent-Guided Co-Scientist for Side-Effect-Aware Drug Redesign}

\begin{document}

\twocolumn[
  \icmltitle{A Precedent-Guided Co-Scientist for Side-Effect-Aware Drug Redesign}



  \icmlsetsymbol{equal}{*}

\begin{icmlauthorlist}
  \icmlauthor{Yujin Kim}{univ}
  \icmlauthor{Charmgil Hong}{univ}
\end{icmlauthorlist}

\icmlaffiliation{univ}{Department of Computer Science and Electrical Engineering, Handong Global University, Pohang, Republic of Korea}

\icmlcorrespondingauthor{Charmgil Hong}{charmgil@handong.edu}
  \icmlkeywords{Machine Learning, ICML}

  \vskip 0.3in
]



\printAffiliationsAndNotice{}  % no special notice (required even if empty)

\begin{abstract}
We propose PRECEDE, a precedent-guided co-scientist for side-effect-aware drug redesign that revises a parent compound to mitigate a specified side effect while preserving therapeutic function.
Rather than isolated molecular generation, PRECEDE frames redesign as evidence-grounded reasoning over drug--side-effect associations, biomedical knowledge graphs, and precedents of safety-driven optimization, coordinated by an LLM orchestrator with explicit policies and human-review checkpoints.
We position PRECEDE as a human-supervised AI-for-science workflow in which hypotheses remain auditable, falsifiable, and bounded by prior pharmacology.
% 
% We propose PRECEDE, a precedent-guided co-scientist for side-effect-aware drug redesign. PRECEDE aims to reduce a specified adverse effect of a parent compound while preserving therapeutic function. We frame this task as evidence-grounded reasoning rather than isolated molecular generation. The system uses drug--side effect evidence, biomedical knowledge graphs, and structured redesign precedents to identify transferable modification strategies. An LLM-based orchestrator coordinates evidence retrieval, constrained redesign, \emph{in silico} docking and toxicity assessment, and iterative refinement. We evaluate PRECEDE through evidence grounding, precedent retrieval quality, redesign quality, efficacy--safety trade-offs, and historical replay against known optimization cases. This proposal positions side-effect mitigation as a human-supervised AI-for-science workflow for explainable molecular redesign.
\end{abstract}

% \vspace{-2 em}
\section{Motivation and Task Definition}

Many drug candidates exhibit therapeutic activity yet carry side effects that require mitigation or restrict clinical use.
Existing molecular optimization methods mainly target generic objectives, such as potency, drug-likeness, or broad ADMET properties~\cite{jin2018junction,you2018graph}.
They typically treat each optimization instance as an independent generation problem. 
Pharmacological redesign, in contrast, often follows recognizable precedents, as in the transition from \textit{tenofovir disoproxil fumarate} to \textit{tenofovir alafenamide}~\cite{ray2016tenofovir}.
In these cases, safety liabilities are mitigated through interpretable structural or pharmacokinetic strategies, such as prodrug modification, bioisosteric replacement, and exposure modulation~\cite{rautio2018expanding,kim2024deepbioisostere}.
Recent agentic drug-discovery systems demonstrate the potential of LLM-based orchestration, retrieval, and tool use for molecular design~\cite{averly2025liddia,liu2024drugagent}. 
% PRECEDE complements this line of work by grounding redesign in prior safety-driven evidence rather than property-score optimization.
% Prior agentic
These systems search chemical space toward a target and a property profile. 
PRECEDE instead takes an approved drug and one documented side effect as input, classifies the effect by mechanism before any edit, and transfers a strategy abstracted from redesign precedents.
% Appendix~\ref{app:positioning} details this positioning.

We define \textit{side-effect-aware drug redesign} as follows.
Given a parent drug $x$, a target side effect $a$, and an optional therapeutic context $c$, the system produces ranked candidates $\{x'_i\}$ along with evidence trails, predicted efficacy--safety profiles, modification rationales, and an auditable redesign report.
Unlike generic molecular generation, the task requires mechanism-informed screening of the side effect before any structural edit: the system must corroborate the drug--side-effect association, identify plausible liability factors, retrieve analogous precedents, and check whether candidate edits preserve the therapeutic mechanism.

% requires scientific decisions before generation as well as after generation. The system must verify the drug--side-effect association, identify plausible biological or structural liability factors, retrieve analogous redesign precedents, evaluate whether candidate modifications preserve the relevant therapeutic mechanism.

PRECEDE addresses this task as a precedent-guided agentic workflow rather than a single generative model.
It corroborates drug--side-effect associations using SIDER~\cite{kuhn2016sider} and knowledge graphs such as PrimeKG~\cite{chandak2023building}, classifies the side effect by likely mechanism, retrieves analogous precedents, and evaluates candidate edits with \emph{in silico} efficacy and safety proxies.
% It verifies drug--side-effect associations with resources such as SIDER~\cite{kuhn2016sider} and biomedical knowledge graphs such as PrimeKG~\cite{chandak2023building}, classifies the adverse effect by likely mechanism, retrieves analogous precedents, and evaluates candidate edits with \emph{in silico} efficacy and safety proxies.
The hypothesis stage, from evidence grounding through \emph{in silico} evaluation and reporting, is handled by the system, while experimental execution and final interpretation remain with human collaborators.
This division keeps outputs as auditable hypotheses rather than autonomous recommendations.

This proposal contributes (i) a precedent memory that encodes historical redesign cases as structured records rather than isolated molecules, (ii) an attribution-aware decision policy that gates structural editing on the redesignability of the side effect, and (iii) a historical replay protocol that tests whether the reasoning of PRECEDE aligns with documented optimization decisions.

% This proposal makes three contributions. First, it formulates side-effect-aware drug redesign as an evidence-grounded AI-for-science task. Second, it introduces a precedent memory that represents historical redesign examples as structured case records rather than isolated molecules. Third, it proposes a human-supervised agentic workflow with a historical replay protocol to assess whether redesign hypotheses reflect prior evidence plus known optimization logic.

\begin{figure*}[t]
    \centering
    \includegraphics[width=\textwidth]{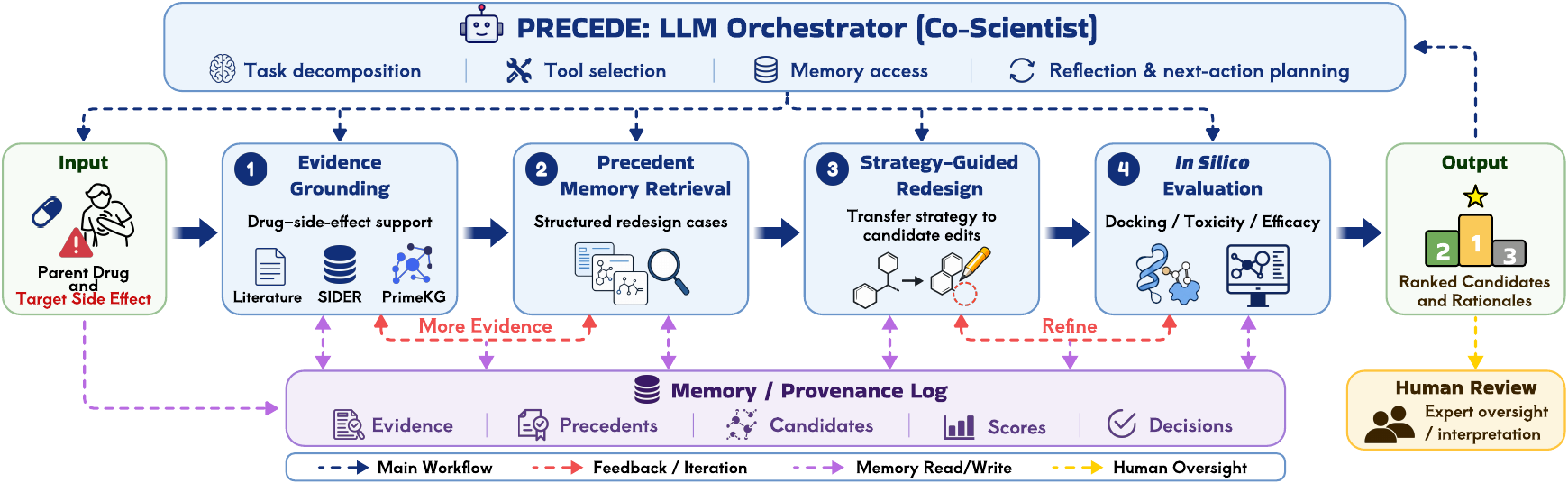}
    \caption{Overview of PRECEDE, an LLM-orchestrated workflow for evidence-grounded, precedent-guided drug redesign.}
    % \caption{Overview of PRECEDE. An LLM-based orchestrator coordinates evidence grounding, precedent retrieval, redesign, and \emph{in silico} evaluation with feedback loops, memory logging, and human oversight.}
    \label{fig:workflow}
\end{figure*}

\section{Proposed Approach}
\label{sec:approach}

PRECEDE uses an LLM-based orchestrator to coordinate evidence retrieval, attribution, precedent search, constrained redesign, and candidate evaluation.
Given a parent compound, a target side effect, and an optional therapeutic context, the orchestrator first checks whether external evidence supports the queried drug--side-effect association.
% Given a parent compound, target adverse effect, and optional therapeutic context, the orchestrator first verifies external support for the queried drug--side-effect association from heterogeneous medicinal-chemistry sources (Section~\ref{sec:eval}).

\textbf{Attribution-aware routing.}\hspace{.25em}
% Before redesign, PRECEDE classifies the side effect into one of five attribution categories based on established mechanism-based side-effect taxonomies~\cite{edwards2000adverse,aronson2003joining,park2011managing}: (i) target-mediated, (ii) off-target structural liability, (iii) metabolism or reactive intermediate, (iv) exposure or pharmacokinetic, and (v) insufficient evidence.
% Only categories (ii)--(iv) proceed to structural redesign; categories (i) and (v) are routed to human review.
% This routing makes structural editing scientifically defensible and constrains the search to redesignable liabilities.
Before redesign, PRECEDE classifies the side effect into one of five categories based on established mechanism-based taxonomies~\cite{edwards2000adverse,aronson2003joining,park2011managing}: (i) target-mediated, (ii) off-target structural liability, (iii) metabolism or reactive intermediate, (iv) exposure or pharmacokinetic, and (v) insufficient evidence.
Only categories (ii)--(iv) proceed to redesign; categories (i) and (v) are routed to human review.
This routing keeps structural editing defensible and constrained to redesignable liabilities.

\textbf{Precedent memory and strategy abstraction.}\hspace{.25em}
Each precedent record encodes a parent compound, safety issue, structural modification, improved compound, and observed outcome.
PRECEDE abstracts transferable strategies from these records, such as exposure modulation, liability-group replacement, pharmacophore preservation, and peripheral editing.
The orchestrator selects strategies with precedents matching the attribution category and therapeutic context.

\textbf{Constrained redesign and evaluation.}\hspace{.25em}
% Strategies guide a hybrid generation stack: matched molecular pair transformations from precedent memory, template-based bioisosteric replacements~\cite{kim2024deepbioisostere}, and LLM-proposed substitutions, filtered by chemical validity and synthetic accessibility.
Strategies guide a hybrid generation stack: matched molecular pair transformations from precedent memory, template-based bioisosteric replacements~\cite{kim2024deepbioisostere}, and LLM-proposed substitutions. Candidate edits are filtered by chemical validity and synthetic accessibility.
Surviving candidates are evaluated through target-specific docking, pharmacophore retention, toxicity predictors, and parent-similarity constraints.

\textbf{Decision policy.}\hspace{.25em}
% Explicit policies govern transitions between stages: evidence below a confidence threshold (\textit{e.g.}, absence of SIDER-confirmed association or fewer than two supporting citations) triggers retrieval expansion; conflicting precedents are resolved by ranked alternative hypotheses scored on contextual similarity; failed evaluation (\textit{e.g.}, docking degradation beyond a parent-relative margin or low synthetic accessibility) returns the system to redesign with tightened constraints.
Policies govern stage transitions: evidence below a confidence threshold (\textit{e.g.}, no SIDER-confirmed association or fewer than two citations) triggers retrieval expansion; conflicting precedents are resolved by ranked alternatives scored on contextual similarity; failed evaluation (\textit{e.g.}, docking degradation beyond a parent-relative margin or low synthetic accessibility) returns the system to redesign under tighter constraints.
A provenance log records every evidence item, precedent, edit, score, and decision rationale; %, and exports an auditable redesign report at termination.
 at termination, it exports an auditable redesign report.

% , then retrieves historical redesign precedents by molecular similarity, side-effect category, therapeutic context, or modification strategy. Candidate sources include curated medicinal chemistry reports, matched molecular pair analyses, drug-label evidence, and literature-mined optimization trajectories.

% The core component is a structured precedent memory. Each precedent record contains a parent compound, safety issue, structural modification, improved compound, and observed outcome. PRECEDE abstracts transferable strategies from these records, such as exposure modulation, liability-group replacement, pharmacophore preservation, or peripheral structure editing. These strategies guide constrained molecular edits, which are evaluated with \emph{in silico} proxies, including target-specific docking, pharmacophore retention, toxicity predictors, and synthetic accessibility.

% The framework is agentic because intermediate outcomes alter the next action. Weak evidence triggers additional retrieval. Uncertain attribution leads to alternative hypotheses. Failed evaluation returns the system to redesign with revised constraints. A memory and provenance log stores evidence, precedents, candidate edits, evaluation scores, and failure reasons to support iterative refinement and auditability.

\section{Evaluation Roadmap}
\label{sec:eval}

PRECEDE is evaluated along four axes that follow its workflow. \emph{Evidence grounding} is assessed by support recall@$k$ against SIDER and PrimeKG, by citation validity, and by attribution classification accuracy on a labeled subset of cases.
\emph{Precedent retrieval} is measured by top-$k$ accuracy and strategy-level F1 on held-out cases. %, distinguishing molecular-similarity recall from the harder question of strategy identification.
\emph{Redesign quality} combines chemical validity, parent similarity, synthetic accessibility, predicted toxicity reduction, and therapeutic preservation via docking and pharmacophore retention as proxies where assay data are unavailable.
\emph{Expert review} provides a multi-rater check on rationale plausibility.

The main benchmark is historical replay on a pilot of 30--50 trajectories from prodrug case studies~\cite{rautio2018expanding} and matched molecular pair literature, each pairing a parent  with its safety-driven successor.
% Holding out the successor, PRECEDE must recover the documented optimization from the parent and target side effect alone.
Holding out the successor makes each hypothesis falsifiable against a documented outcome: PRECEDE must recover the optimization from the parent and target side effect alone.
We treat replay as necessary but not sufficient: agreement supports medicinal-chemistry reasoning, while interpretable disagreement remains informative.
A proof-of-concept implementation and single-case walkthrough are in the appendix.

\section{Governance and Expected Impact}
\label{sec:gov}
% PRECEDE is positioned as a human-supervised hypothesis-generation system rather than an autonomous clinical or synthesis engine: it neither recommends clinical use nor initiates synthesis; its outputs are computational hypotheses of open causal status.
% %This stance is enforced by the attribution-aware routing of Section~\ref{sec:approach}.
% Human-review is invoked at three checkpoints (non-redesignable or insufficient-evidence cases, conflicting precedents, and final candidate triage).
% Every evidence item, precedent, edit, score, and human intervention is logged to the provenance memory for attribution and audit.
% The aim is not to automate medicinal chemistry but to provide a testbed in which the reasoning behind safety-driven redesign remains legible, contestable, and accountable to humans who remain responsible for the science.

PRECEDE is positioned as a co-scientist: a human-supervised hypothesis-generation system rather than an autonomous clinical or synthesis engine.
It proposes evidence-grounded redesign hypotheses, while humans retain decision authority and scientific credit.
It neither recommends clinical use nor initiates synthesis; its outputs are computational hypotheses of open causal status.
Human review is invoked at three checkpoints (non-redesignable or insufficient-evidence cases, conflicting precedents, and final candidate triage); every evidence item, precedent, edit, score, and human intervention is logged to the provenance memory.
The aim is not to automate medicinal chemistry but to offer medicinal chemists an auditable testbed in which safety-driven redesign remains legible, contestable, and accountable, extensible to other precedent-guided design tasks.

% PRECEDE provides a testbed for agentic scientific workflows that bridge molecular design and translational decision-making. By grounding redesign in prior cases and explicit evidence, the system may support early lead optimization where safety liabilities must be reduced without loss of therapeutic function.

% PRECEDE is designed as a human-supervised hypothesis-generation system, not as an autonomous clinical or synthesis engine. It does not recommend clinical use, initiate synthesis, or make treatment decisions. Because adverse effects may arise from target biology, off-target activity, metabolism, exposure, dose, or patient-specific factors, redesign suggestions are treated as computational hypotheses rather than causal conclusions. To support transparency, the memory log records evidence sources, retrieved precedents, candidate modifications, evaluation scores, and decision rationales. Low-confidence or high-risk outputs are flagged for human review and experimental validation.

\section*{Acknowledgements}
% This research was supported by the Regional Innovation System \& Education(RISE) program through the Gyeongbuk RISE Center, funded by the Ministry of Education(MOE) and the Gyeongsangbuk-do, Republic of Korea. (2026-RISE-15-119)
% This research was supported by the MSIT (Ministry of Science, ICT), Korea, under the Global Research Support Program in the Digital Field program (RS-2024-00431394) supervised by the IITP (Institute for Information \& Communications Technology Planning \& Evaluation).
This work was supported by the Ministry of Science and ICT (MSIT), Republic of Korea, under the Global Research Support Program in the Digital Field (RS-2024-00431394), supervised by the Institute for Information \& Communications Technology Planning \& Evaluation (IITP), and by MSIT under the Advanced GPU Utilization Support Program (05-26-04-0020).


\bibliography{references}
\bibliographystyle{icml2026}

\appendix
\section{Appendix}

\begin{table*}[!b]
\centering
\small
\caption{Evaluation axes, metrics, data sources, and target criteria for the PRECEDE pilot study.}
\label{tab:eval}

\renewcommand{\arraystretch}{1.12}
\setlength{\tabcolsep}{5pt}

\begin{tabularx}{\textwidth}{
@{}
>{\raggedright\arraybackslash}p{2.5cm}
>{\raggedright\arraybackslash}X
>{\raggedright\arraybackslash}p{3.5cm}
>{\raggedright\arraybackslash}X
@{}
}
\toprule
\textbf{Axis} & \textbf{Metric} & \textbf{Source} & \textbf{Target} \\
\midrule
Evidence grounding 
& support recall@$k$, citation validity, attribution accuracy 
& SIDER, PrimeKG, labeled subset 
& high agreement with curated associations \\

Precedent retrieval 
& top-$k$ accuracy, strategy-level F1 
& held-out redesign cases 
& exceeds similarity-only baselines \\

Redesign quality 
& validity, SA, parent Tanimoto, docking $\Delta$, toxicity-proxy $\Delta$ 
& RDKit, AutoDock Vina, ADMET predictors 
& reduced toxicity proxy with preserved efficacy proxy \\

Expert review 
& rationale plausibility (1--5), blinded, multi-rater 
& held-out redesign cases 
& mean $\geq 3.5$, substantial agreement \\
\bottomrule
\end{tabularx}
\end{table*}

\subsection{Positioning Relative to Prior Agentic Systems}
\label{app:positioning}
 
Recent agentic drug-discovery systems share one execution pattern: an LLM orchestrator retrieves data, invokes generators, and filters candidates with \emph{in silico} tools. DrugAgent~\cite{liu2024drugagent} automates the machine-learning programming behind discovery tasks. LIDDIA~\cite{averly2025liddia} coordinates the \emph{in silico} stages of hit identification and lead optimization. Both frame the objective as a search over chemical space toward a target and a property profile. PRECEDE reuses this orchestration pattern, yet its contribution rests on three departures from prior work.
 
\textbf{Problem formulation.} Prior systems take a target and a property specification as input and return molecules that bind the target. PRECEDE takes an approved drug and one documented side effect as input and returns revisions that remove the liability while the therapeutic mechanism stays intact. The task is safety-driven redesign of an existing drug rather than \emph{de novo} generation.
 
\textbf{Precedent-guided reasoning.} These systems optimize property scores for each case in isolation. PRECEDE abstracts a strategy from structured records of how medicinal chemists resolved analogous liabilities and transfers that strategy by analogy. The precedent memory is a resource that property-score optimizers do not use.
 
\textbf{Attribution before generation.} Many prior systems follow a generate-and-filter pattern, in which candidate molecules are proposed before downstream property and safety filters are applied. PRECEDE classifies the side effect by mechanism before any structural edit and admits only redesignable liabilities to redesign. Target-mediated effects halt at human review. The gate decides whether redesign is scientifically defensible before the generators run.
 
The contribution rests on the grounds of the reasoning rather than on the orchestration mechanics. The external tools are invoked under fixed execution settings, while the agentic component lies in non-trivial decisions: attribution classification, precedent selection under conflict, the reflect--refine--escalate loop, and rationale generation. Each decision is grounded in external evidence and recorded in the provenance log. The exception cases in Appendix~\ref{app:walkthrough} (warfarin and terfenadine) demonstrate that the pipeline branches and rejects inputs rather than executes a fixed batch.

\subsection{Pilot Evaluation Protocol}
\label{app:eval}

Table~\ref{tab:eval} summarizes the four evaluation axes introduced in Section~\ref{sec:eval}. It specifies the metrics, data sources, and target criteria used to operationalize evidence grounding, precedent retrieval, redesign quality, and expert review in the initial PRECEDE pilot.

\textbf{Support recall@$k$} measures the fraction of held-out drug--side-effect associations for which relevant supporting evidence appears in the top-$k$ results. \textbf{Citation validity} measures the fraction of cited sources that resolve to a verifiable record in SIDER, PrimeKG, PubMed, or the corresponding source database. \textbf{Attribution accuracy} is computed on a manually labeled subset of cases with literature-documented side-effect mechanisms; it tests whether PRECEDE assigns each case to the correct category among the five categories in Section~\ref{sec:approach}.
\textbf{Strategy-level F1} maps each retrieved precedent to an abstracted strategy class (Section~\ref{sec:approach}) and compares it with the strategy documented in the held-out case. 
\textbf{Toxicity-proxy $\Delta$} measures the relative change in predicted toxicity between parent and candidate compounds, averaged across multiple ADMET predictors to reduce dependence on a single model. 
\textbf{Expert review} uses three independent raters to score rationale plausibility on a 1--5 Likert scale. Inter-rater agreement is reported using Cohen's $\kappa$ or an appropriate multi-rater agreement statistic.

\textbf{Candidate success criterion.} A candidate counts as a successful redesign only when it satisfies three conditions jointly. First, the therapeutic efficacy proxy stays within the parent-relative margin. Second, the targeted safety endpoint improves relative to the parent. Third, no non-targeted safety endpoint degrades beyond the parent-relative margin. The third condition screens the full ADMET-AI~\cite{swanson2024admet} panel to flag cases where a redesign may trade the targeted liability for a new one.

\begin{table*}[t]
\centering
\footnotesize
\caption{Seed precedent memory used by PRECEDE in the current proof-of-concept implementation.}
\label{tab:precedents}
\renewcommand{\arraystretch}{1.15}
\setlength{\tabcolsep}{4pt}

\begin{tabularx}{\textwidth}{
@{}
>{\raggedright\arraybackslash}p{3.2cm}
>{\raggedright\arraybackslash}p{3.2cm}
>{\raggedright\arraybackslash}X
>{\raggedright\arraybackslash}p{2.6cm}
@{}
}
\toprule
\textbf{Parent $\rightarrow$ successor} &
\textbf{Targeted side effect} &
\textbf{Strategy and structural change} &
\textbf{Attribution} \\
\midrule
TDF $\rightarrow$ TAF &
nephrotoxicity; bone mineral density loss &
exposure modulation via phosphonamidate prodrug conversion &
Exposure / PK \\

cisplatin $\rightarrow$ carboplatin &
nephrotoxicity &
liability-group replacement via cyclobutanedicarboxylate leaving group &
Exposure / PK \\

carbamazepine $\rightarrow$ oxcarbazepine &
hypersensitivity (rash) &
keto-analog strategy via 10-keto substitution &
Metabolism / reactive \\

terfenadine $\rightarrow$ fexofenadine &
QT prolongation; torsades (hERG block) &
active-metabolite switch via carboxylic-acid active metabolite &
Off-target structural \\
\bottomrule
\end{tabularx}
\end{table*}

\begin{table}[t]
\centering
\small
\caption{Knowledge resources and computational tools used across task modules in the implemented pipeline.}
\label{tab:stack}
\begin{tabularx}{\columnwidth}{
@{}
>{\raggedright\arraybackslash}p{3.1cm}
>{\raggedright\arraybackslash}X
@{}
}
\toprule
\textbf{Task module} & \textbf{Knowledge resources and tools} \\
\midrule
Evidence grounding & SIDER, OnSIDES \\
Target resolution & ChEMBL, UniProt, RCSB PDB \\
SMILES generation & mmpdb, REINVENT Mol2Mol \\
Property prediction & ADMET-AI, RDKit \\
Binding / efficacy & AutoDock Vina \\
\bottomrule
\end{tabularx}
\end{table}

\subsection{Pilot Benchmark Curation}
\label{app:benchmark}

The historical replay benchmark is designed around 30--50 safety-driven redesign trajectories from two complementary sources. The first is curated medicinal chemistry case studies, including prodrug conversion and bioisosteric replacement~\cite{rautio2018expanding,kim2024deepbioisostere}. An example is the transition from tenofovir disoproxil fumarate to tenofovir alafenamide.
The second source is matched molecular pair literature derived from public bioactivity databases such as ChEMBL~\cite{gaulton2012chembl}, where structural transformations associated with reduced off-target activity, such as hERG or CYP inhibition, have been systematically characterized~\cite{kramer2014matched}. Each trajectory contains a parent compound, a documented side effect, a structural modification, and a successor compound, with the modification rationale recorded as a ground-truth strategy label.

The pilot is kept deliberately modest in scale because the value of historical replay depends on the verifiability of each trajectory rather than the size of the corpus. Trajectories with ambiguous attribution, undocumented design rationale, or insufficient outcome data are excluded. We anticipate that scaling beyond the pilot will require automated mining of medicinal chemistry literature, which we treat as future work conditional on the validation of the pilot protocol.

\subsection{Prototype Implementation Details}
\label{app:implementation}
 
PRECEDE runs on a LangGraph orchestration graph~\cite{Chase_LangChain_2022} with conditional edges and a bounded reflection loop. The LLM is restricted to attribution, reflection, and rationale generation, while external tools perform retrieval, generation, scoring, and filtering under fixed execution settings. The backbone is Qwen3-30B-A3B-FP8~\cite{yang2025qwen3}. Table~\ref{tab:stack} lists the tool components per task module. Table~\ref{tab:precedents} lists the seed precedent memory used in the current proof-of-concept implementation.

\subsection{Prototype Workflow Walkthrough}
\label{app:walkthrough}
This section traces the implemented pipeline on a single case that the precedent memory does not contain, to demonstrate the control flow of PRECEDE rather than to validate the biological validity of the generated analog.
The case separates strategy transfer from retrieval of a documented successor. 
The parent drug is cidofovir, a nucleotide analog with well-documented renal toxicity, and the target side effect is renal failure. 
The precedent memory holds no successor of cidofovir; the case therefore tests transfer of an abstracted strategy rather than retrieval of a stored answer.

\textbf{Evidence grounding.}
The evidence module queries the side-effect resources SIDER and OnSIDES~\cite{tanaka2025onsides}.
Both resources support the cidofovir--renal failure association through two independent sources, with the matched terms ``Renal failure'' and ``Renal failure acute''.
The association meets the confidence threshold; the case proceeds to attribution.

\textbf{Attribution-aware routing.}
The orchestrator assigns the exposure or pharmacokinetic (PK) category among the five categories of Section~\ref{sec:approach}. 
The recorded rationale attributes the toxicity to renal tubular accumulation rather than to an on-target or off-target interaction. 
The PK category is redesignable; the case advances to precedent retrieval.

\textbf{Precedent retrieval and strategy transfer.} 
The retrieval module ranks precedents by attribution-category match and by drug and effect token overlap. 
The two top precedents are the transition from tenofovir disoproxil fumarate to tenofovir alafenamide (TDF $\rightarrow$ TAF) and from cisplatin to carboplatin. 
Both precedents share the PK category with the query; neither precedent is a documented redesign of cidofovir. 
The module abstracts the strategy \emph{exposure modulation} from TDF $\rightarrow$ TAF and transfers it to cidofovir.

\textbf{Strategy-guided redesign.}
The redesign module generates candidates from the parent structure with two generators: matched molecular pair (MMP)~\cite{hussain2010computationally} transformations and a deep generative model (REINVENT Mol2Mol)~\cite{loeffler2024reinvent}. 
The module retains $23$ valid candidates, $17$ from REINVENT Mol2Mol and $6$ from MMP transformations, after a chemical-validity filter and the removal of duplicates and parent-identical structures.
 
\begin{figure}[t]
    \centering
    \includegraphics[width=0.48\textwidth]{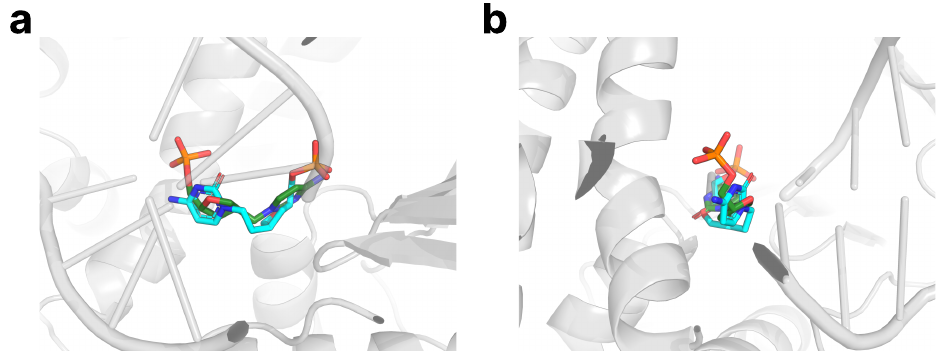}
    \caption{Docked poses on viral DNA polymerase (PDB 9JA3). Green: cidofovir (parent); cyan: generated analog.}
    \label{fig:docking}
\end{figure}

\textbf{\textit{In silico} evaluation.}
The evaluation module resolves a therapeutic target of cidofovir to the herpes simplex virus type 1 (HSV-1) DNA polymerase (UniProt accession P04293)~\cite{uniprot2018uniprot} and selects Protein Data Bank (PDB) structure 9JA3~\cite{burley2021rcsb}, a complex of the HSV-1 polymerase with duplex DNA and a bound nucleotide-analog inhibitor, for docking. 
The module predicts toxicity with ADMET-AI and selects microsomal intrinsic clearance (\texttt{Clearance\_Microsome\_AZ}) as the primary safety endpoint for the PK category. 
A higher clearance value indicates lower systemic exposure; higher values are treated as more favorable for this endpoint.
The module ranks the $23$ candidates by the docking-independent score components (safety
endpoint delta, parent Tanimoto similarity, and synthetic accessibility). 
It then docks the parent and the three top-ranked candidates with AutoDock Vina~\cite{trott2010autodock} (Figure~\ref{fig:docking}).
The parent docking score is $-7.88$~kcal/mol. 
For these three docked candidates, docking-score deltas span $-0.30$ to $+0.24$~kcal/mol; all three remain within the parent-relative efficacy margin of $2.0$~kcal/mol.
The reflection module evaluates the top candidate and returns an accept decision (efficacy preserved, scaffold similarity maintained), terminating the loop.

\begin{table}[t]
\centering
\small
\caption{Rank-1 candidate for the cidofovir walkthrough. The candidate originates
from a strategy transferred from the TDF $\rightarrow$ TAF precedent.}
\label{tab:cidofovir}
\begin{tabular}{ll}
\toprule
\textbf{Quantity} & \textbf{Value} \\
\midrule
Attribution category & exposure or pharmacokinetic \\
Transferred strategy & exposure modulation \\
% Candidate source & REINVENT Mol2Mol \\
Parent Tanimoto similarity & $0.83$ \\
Primary safety endpoint & \texttt{Clearance\_Microsome\_AZ} \\
Safety endpoint delta & $+3.65$ (improvement) \\
Docking score (parent) & $-7.88$~kcal/mol \\
Docking score (candidate) & $-8.18$~kcal/mol \\
Efficacy preserved & yes \\
Composite score & $0.94$ \\
\bottomrule
\end{tabular}
\end{table}

% \begin{figure}[t]
%     \centering
%     \includegraphics[width=0.48\textwidth]{Figure/candidate_barplot.png}
%     \caption{Per-candidate evaluation for the cidofovir walkthrough. Top: docking score change relative to the parent for the docked candidates ($\Delta$ vs.\ parent, kcal/mol); lower values indicate stronger predicted binding; the dashed line marks the $2.0$~kcal/mol efficacy margin. Bottom: predicted change in the primary safety endpoint (\texttt{Clearance\_Microsome\_AZ}) relative to the parent; higher values indicate lower systemic exposure.}
%     \label{fig:barplot}
% \end{figure}

\textbf{Ranking and report.}
The system ranks candidates with a transparent weighted score over four normalized components: the primary safety endpoint delta (weight $0.45$), docking efficacy (weight $0.30$), parent Tanimoto similarity (weight $0.15$), and synthetic accessibility (SA, weight $0.10$). 
The score is normalized over the components available for each candidate. Table~\ref{tab:cidofovir} summarizes the rank-1 candidate, which originates from REINVENT Mol2Mol, has a parent Tanimoto similarity of $0.83$, and reaches a composite score of $0.94$.
The candidate docking score is $-8.18$~kcal/mol, a delta of $-0.30$ relative to the parent. The generated rationale frames the candidate as consistent with an exposure-modulation hypothesis derived from the transferred precedent strategy. 
Microsomal clearance is a hepatic measure and serves as an indirect proxy for renal exposure; this limitation motivates the human-supervised positioning of the system. 
The workflow terminates with a final human-review checkpoint for expert triage that marks every candidate as a hypothesis rather than a clinical or synthesis recommendation.

\textbf{Exception cases.}
Two cases test the first human-review checkpoint, where non-redesignable or insufficient-evidence inputs should halt before structural editing.
The first case pairs warfarin with bleeding. The orchestrator assigns the target-mediated category because bleeding extends the intended anticoagulant mechanism rather than an off-target or pharmacokinetic liability. Because the target-mediated category is non-redesignable, the case halts at a human-review checkpoint without structural editing.
The second case pairs terfenadine with QT prolongation. Although terfenadine appears in the seed precedent memory, the evidence module does not allow it to substitute for external side-effect support. Because the queried association is not covered by SIDER or OnSIDES in the current implementation, the case is assigned to the insufficient-evidence category and halts at a human-review checkpoint.
Both cases illustrate that evidence grounding and attribution routing gate the pipeline as specified. Neither query reaches redesign.

\subsection{Governance and Risk Considerations}
\label{app:governance}

\textbf{Data provenance.} %All evidence sources used in PRECEDE (SIDER, PrimeKG, ChEMBL-derived MMP datasets, published case studies) are publicly available and licensed for research use. The system does not ingest patient-level or proprietary data.
PRECEDE relies on public research resources, including SIDER, PrimeKG, ChEMBL-derived matched molecular pair datasets, and published medicinal chemistry case studies. The system does not ingest patient-level data or proprietary datasets in the proposed pilot.

\textbf{Dual-use considerations.} Side-effect mitigation strategies could, in principle, inform the inverse problem of increased side-effect risk. PRECEDE mitigates this risk by restricting outputs to literature-grounded modifications and abstracted strategies, and by logging generated candidates to the provenance memory. The system uses REINVENT only in parent-conditioned Mol2Mol mode and does not perform \emph{de novo} toxicophore exploration or autonomous synthesis planning.

\textbf{Human oversight protocol.} The three review checkpoints described in Section~\ref{sec:gov} are operationalized as follows. %Cases routed to non-redesignable attribution categories are presented to a human reviewer with the supporting evidence and a brief justification of the routing decision. Conflicting precedents above a disagreement threshold trigger a human triage step in which alternative hypotheses and their contextual similarity scores are surfaced for selection. Final candidate triage involves expert review of ranked candidates against the auditable redesign report before any downstream use.
First, cases routed to non-redesignable or insufficient-evidence categories are presented to a human reviewer with the supporting evidence and routing justification.
Second, conflicting precedents above a disagreement threshold trigger human triage, where alternative hypotheses and contextual similarity scores are surfaced for selection.
Third, candidate triage requires expert review of ranked candidates against the auditable redesign report before any downstream use.

\textbf{Limitations.} %PRECEDE relies on the completeness and accuracy of public adverse-effect databases, which are known to under-report long-tail effects and to encode reporting biases. Predicted toxicity proxies are imperfect and can exhibit domain shift on novel scaffolds. Historical replay agreement is consistent with documented logic but does not establish causal validity for novel cases. These limitations motivate the human-supervised positioning of the system rather than autonomous deployment.
PRECEDE depends on public side-effect databases, which can under-report long-tail effects and encode reporting biases.
Predicted toxicity proxies may exhibit domain shift on novel scaffolds.
Historical replay agreement indicates consistency with documented medicinal chemistry logic, but it does not establish causal validity for novel redesign cases.
These limitations motivate the human-supervised positioning of PRECEDE rather than autonomous deployment.
%%%%%%%%%%%%%%%%%%%%%%%%%%%%%%%%%%%%%%%%%%%%%%%%%%%%%%%%%%%%%%%%%%%%%%%%%%%%%%%
%%%%%%%%%%%%%%%%%%%%%%%%%%%%%%%%%%%%%%%%%%%%%%%%%%%%%%%%%%%%%%%%%%%%%%%%%%%%%%%

\end{document}